\documentclass[letterpaper]{article} 
\usepackage{aaai23}
\usepackage{times}  
\usepackage{helvet}  
\usepackage{courier}  
\usepackage[hyphens]{url}  
\usepackage{graphicx} 
\usepackage{natbib}  
\usepackage{caption} 
\usepackage[super]{nth}
\usepackage{amsmath}
\usepackage{bm}
\usepackage{array}
\usepackage{makecell}
\usepackage{multirow}
\usepackage{tabularx}
\usepackage{graphicx}
\usepackage{caption}
\usepackage{diagbox}
\usepackage{subcaption}
\usepackage{url}
\usepackage{balance}
\usepackage{enumitem}
\usepackage{float}
\usepackage[flushleft]{threeparttable}
\usepackage{bbm}
\usepackage{graphicx} 
\usepackage{xcolor}
\usepackage{amsfonts}
\usepackage{hyperref}
\hypersetup{
    colorlinks=true,
    linkcolor=black,
    filecolor=black,      
    citecolor=black,
    urlcolor=blue,
    }

\usepackage{etoolbox}
\usepackage{environ}

\newtoggle{comments}
 \toggletrue{comments} 
\NewEnviron{doweprint}[1]{%
  \iftoggle{#1}{\BODY}{}%
}

\usepackage[ruled,vlined,linesnumbered]{algorithm2e}

\newcommand*\dif{\mathop{}\!\mathrm{d}}

\usepackage{newfloat}
\usepackage{listings}
\DeclareCaptionStyle{ruled}{labelfont=normalfont,labelsep=colon,strut=off} 
\floatstyle{ruled}
\newfloat{listing}{tb}{lst}{}
\floatname{listing}{Listing}
\title{Fairness and Explainability: Bridging the Gap \\Towards Fair Model Explanations}
\author{
    Yuying Zhao, Yu Wang, Tyler Derr 
}
\affiliations{
    Vanderbilt University 

    \{yuying.zhao, yu.wang.1, tyler.derr\}@vanderbilt.edu 
}

\usepackage{bibentry}

\begin{document}

\maketitle

\begin{abstract}
While machine learning models have achieved unprecedented success in real-world applications, they might make biased/unfair decisions for specific demographic groups and hence result in discriminative outcomes. Although research efforts have been devoted to measuring and mitigating bias, 
they mainly study bias from the result-oriented perspective while neglecting the bias encoded in the decision-making procedure. This results in their inability to capture procedure-oriented bias, which therefore limits the ability to have a fully debiasing method.
Fortunately, with the rapid development of explainable machine learning, explanations for predictions are now available to gain insights into the procedure. In this work, we bridge the gap between fairness and explainability by presenting a novel perspective of procedure-oriented fairness based on explanations. We identify the procedure-based bias by measuring the gap of explanation quality between different groups with Ratio-based and Value-based Explanation Fairness. 
The new metrics further motivate us to design an optimization objective to mitigate the procedure-based bias where we observe that it will also mitigate bias from the prediction. 
Based on our designed optimization objective, we propose a Comprehensive Fairness Algorithm (CFA), which simultaneously fulfills multiple objectives - improving traditional fairness, satisfying explanation fairness, and maintaining the utility performance. 
Extensive experiments on real-world datasets demonstrate the effectiveness of our proposed CFA and highlight the importance of considering fairness from the explainability perspective. 
Our code: \href{https://github.com/YuyingZhao/FairExplanations-CFA}{https://github.com/YuyingZhao/FairExplanations-CFA}.

\end{abstract}

\section{Introduction}\label{sec-introduction}
Recent years have witnessed the unprecedented success of applying machine learning (ML) models in real-world domains, such as improving the efficiency of information retrieval~\cite{recommendation_example, wang2022collaboration} and providing convenience with intelligent language translation~\cite{language_translation}.
However, recent studies have revealed that historical data may include patterns of previous discriminatory decisions dominated by sensitive features such as gender, age, and race~\cite{2017survey-fairness,2021survey-fairness}. Training ML models with data including such historical discrimination would explicitly inherit existing societal bias and further lead to cascading unfair decision-making in real-world applications~\cite{finance_bias, education_bias, healthcare_bias}. 

In order to measure bias of decisions so that further optimization can be provided for mitigation, a wide range of metrics have been proposed to quantify unfairness
~\cite{2012groupfairness,2016post,2017survey-fairness}. Measurements existing 
in the literature can be generally divided into group fairness and individual fairness where group fairness~\cite{2012groupfairness, 2016post} measures the similarity of model predictions among different sensitive groups (e.g., gender, race, and income) and individual fairness~\cite{kusner2017counterfactual} measures the similarity
of model predictions among similar individuals. However, all these measurements are computed based on the outcome of model predictions/decisions while ignoring the procedure leading to such predictions/decisions.
In this way, the potential bias hidden in the decision-making process will be ignored and hence cannot be further mitigated by result-oriented debiasing methods, leading to a restricted mitigation.

Despite the fundamental importance of measuring and mitigating the procedure-oriented bias from explainability perspective, the related studies are still in their infancy~\cite{2020aaai-concern, sigir2020-fair-recommendation}. To fill this crucial gap, we widen the focus from solely measuring and mitigating the result-oriented bias to further identifying and alleviating the procedure-oriented bias. 
However, this is challenging due to the inherent complexity of ML models which aggravates the difficulty in understanding the procedure-oriented bias.  Fortunately, with the development of explainability which aims to demystify the underlying mechanism of why a model makes prediction, we thus have a tool of generating human-understandable explanations for the prediction of a model, and therefore have insights into decision-making process. Given the explanations, a direct comparison would suffer from inability of automation due to the requirement of expert knowledge to understand the relationship between features and bias. The explanation quality, which is domain-agnostic, can serve as an indicator for fairness. If the quality of explanations are different for sensitive groups, the model treats them unfairly presenting higher-quality explanations for one group than the other, which indicates the model is biased. To capture such bias encoded in the decision-making procedure, we bridge the gap between fairness and explainability and provide a novel procedure-oriented perspective based on explanation quality. We propose two group explanation fairness metrics measuring the gap between explanation quality among sensitive groups. First, \underline{R}atio-based \underline{E}xplanation \underline{F}airness $\Delta_{\text{REF}}$ extends from 
result-oriented metric and quantifies the difference between ratios of instances with high-quality explanations for sensitive groups. However, due to the simplification from the continual scores to a discrete binary quality (e.g., high/low), $\Delta_{\text{REF}}$ will ignore certain bias, motivating \underline{V}alue-based \underline{E}xplanation \underline{F}airness $\Delta_{\text{VEF}}$ based on the detailed quality values. These two metrics cooperate to provide the novel explanation fairness perspective.

Traditional fairness measurements quantify bias in the result-oriented manner and our proposed explanation fairness measures quantify bias in the procedure-oriented way. These metrics together present a comprehensive view of fairness. Attempting to improve fairness from multiple perspectives while maintaining utility performance, we further design a Comprehensive Fairness Algorithm (CFA) based on minimizing representation distances between instances from distinct groups. This distance-based optimization is inspired by traditional result-oriented fairness methods~\cite{2022www-edits,2021nifty}.
Different from vanilla models that only aim for higher utility performance where the instance representations solely serve for the utility task, models with fairness consideration add fair constraints on the instance representations so that the instances from different groups are close to each other in the embedding space and thus avoid bias (i.e., the learned representations are task-related and also fair for sensitive groups). Naturally, we extend this idea to seek fair explanation quality in the embedding space which is composed of two parts: (a) the embeddings based on the original input features, and (b) the embeddings based on the partially masked input features according to feature attributions generated from explanations (inspired by explanation fidelity~\cite{robnik2018perturbation-fidelity}). The objective function for explanation fairness thus becomes minimizing the representation distance between subgroups based on the input features and the masked features. Therefore, our learned representations achieve multiple goals: encoding task-related information and being fair for sensitive groups in both the prediction and the procedure.
Our main contributions are three-fold:
\begin{itemize}
\item \textbf{Novel fairness perspectives/metrics}. Merging fairness and explainability domains, we propose explanation-based fairness perspectives and novel metrics ($\Delta_{\text{REF}}$, $\Delta_{\text{VEF}}$) that identify bias as gaps in explanation quality. 
\item \textbf{Comprehensive fairness algorithm}. We design Comprehensive Fairness Algorithm (CFA) to optimize multiple objectives simultaneously. We observe additionally optimizing for explanation fairness is beneficial for improving traditional fairness.
\item \textbf{Experimental evaluation}. We conduct extensive experiments which verify the effectiveness of CFA and demonstrate the importance of considering procedure-oriented bias with explanations.
\end{itemize}
\section{Related Work}\label{sec-relatedwork}
\subsection{Fairness in Machine Learning}\label{sec:related-fairness}
Fairness in ML has raised increasingly interest in recent years~\cite{fairness-survey, dong2022fairness}, as bias is commonly observed in ML models~\cite{bias-survey}
imposing risks to the social development and impacting individual's lives.
Therefore, researchers have devoted much effort into this field.
To measure the bias, various metrics
have been proposed, including the difference of statistical parity ($\Delta_{\text{SP}}$)~\cite{2012groupfairness}, equal opportunity ($\Delta_{\text{EO}}$)~\cite{2016post}, equalized odds~\cite{2016post}, counterfactual fairness ~\cite{kusner2017counterfactual}, etc.
To mitigate bias, researchers design debiasing solutions through pre-processing~\cite{2011preprocessing},
processing~\cite{2021nifty,reductions,reweight},
and post-processing~\cite{2016post}. 
Existing fairness metrics capture the bias from the predictions while ignoring the potential bias in the procedure. The debiasing methods to improve them therefore mitigate the result-oriented bias rather than procedure-oriented bias.

\subsection{Model Explainability}\label{sec:related-explainability}
Explainability techniques~\cite{explainability-survey-2018} seek to demystify black box ML models by providing human-understandable explanations from varying viewpoints on which parts (of the input) are essential for prediction~\cite{2020survey-exp2,2022survey-exp,2020survey-gnnexplaination}. Based on how the important scores are calculated, they are categorized into gradient-based methods~\cite{2017gradcam,2017gradientshap}, perturbation-based methods~\cite{2017neurips-perturbation}, surrogate methods~\cite{graphlime,lime}, and decomposition methods~\cite{2017taylor-decomposition}.
These techniques help identify most salient features for the model decision, which provide insights into the procedure and reasons behind the model decisions.

\subsection{Intersection of Fairness and Explainability}\label{sec:related-fairness-explainability}
Only few work study the intersection between fairness and explainability and most
focus on explaining existing fairness metrics.
They decompose model disparity into feature~\cite{2020-feature-case-study,2020-feature-based-explanation} or path-specific contributions~\cite{2019aaai-path-explanation,2021kdd-path-explanation} 
aiming to understand the origin of bias.
Recently, in~\cite{2020aaai-concern} fairness was studied from explainability where they observed a limitation of using sensitive feature importance to define fairness. Each of these prior works have not explicitly considered explanation fairness.
To the best of our knowledge, only one work~\cite{sigir2020-fair-recommendation} designed such novel metric based on the diversity of explanations for fairness-aware explainable recommendation. However, it is limited to the specific domain.
In comparison, we seek to find a novel perspective on fairness regarding explanation and quantify explanation fairness in a more general form that can be utilized in different domains.

\section{Novel Fairness Perspectives}
\label{sec-concerns}
In this section, we highlight the significance of procedure-oriented explanation fairness.
Based on explanation quality, we propose Ratio-based and Value-based Explanation Fairness ($\Delta_{\text{REF}}$ and $\Delta_{\text{VEF}}$).
For ease of understanding, we begin with the binary case where sensitive features $s\in\{0, 1\}$ and the whole group $\mathcal{G}$ is split into subgroups $\mathcal{G}_0$ and $\mathcal{G}_1$ based on the sensitive attribute (e.g., white and non-white subgroups divided by race) and investigate multiple-sensitive feature scenario. We include a discussion about result-oriented and procedure-oriented fairness in the end of this section. Notations used in the paper are summarized in Appendix~\ref{sec-notation}.

\subsection{Motivating Explanation Quality}
\label{identify-meq}
\begin{figure}[t]
     \centering
\includegraphics[width=0.48\textwidth]{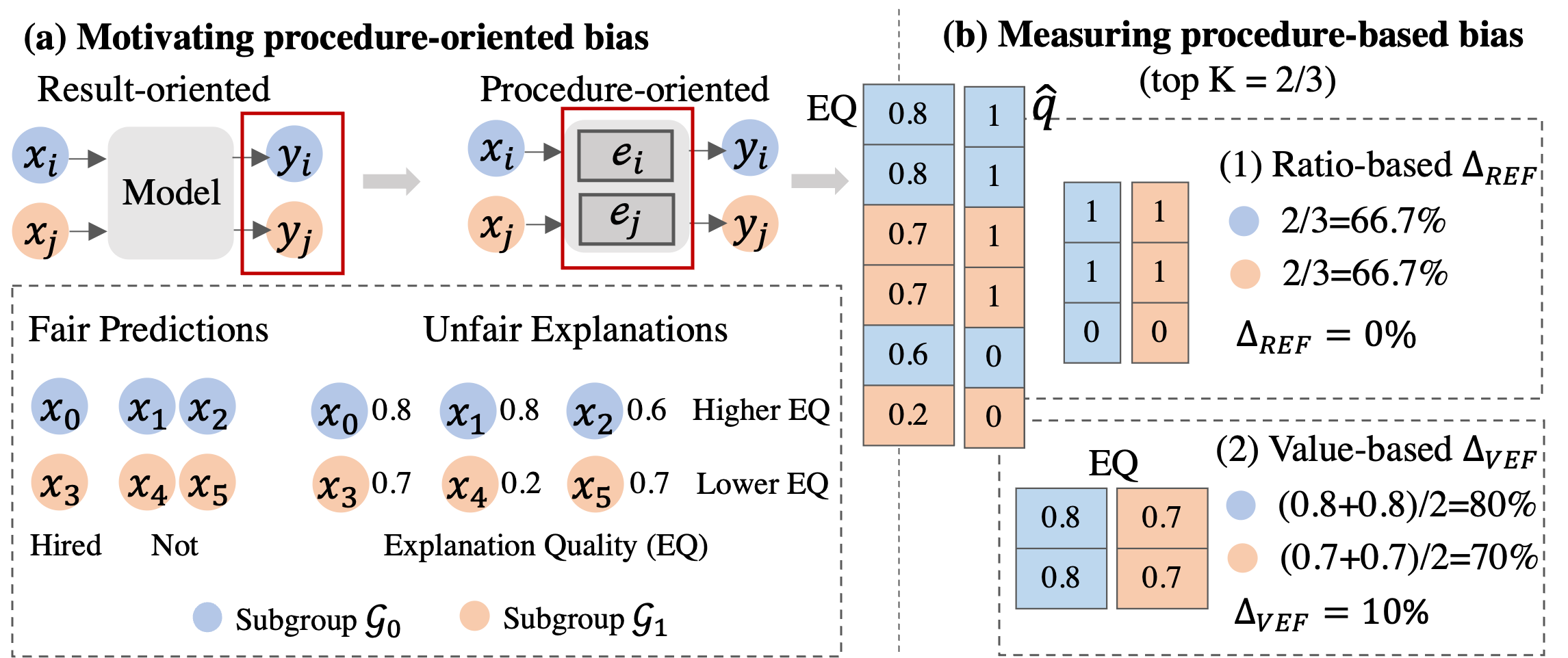}
    \vskip -1.5ex
    \caption{A motivating example of job hiring. From the result-oriented perspective, the prediction is fair since both two subgroups have the same statistical parity. From the procedure-oriented perspective, the explanation is unfair since explanation quality (EQ) of $\mathcal{G}_0$ is generally higher than the one of $\mathcal{G}_1$.}
     \vskip -1.5ex
     \label{fig.motivation}
\end{figure}

Although various fairness metrics have been proposed (Section~\ref{sec:related-fairness}), almost all of them are directly computed based on the outputs of the predictions. Accordingly, the quantified unfairness could only reflect the result-oriented bias while ignoring the potential bias caused by the decision-making procedure. As shown in Figure~\ref{fig.motivation} of a hiring example, model makes hiring decisions based on users' features. From the result-oriented perspective, the model is fair under statistical parity constraint. However, from the procedure-oriented perspective, the model contains implicit bias during the decision-making process.
The neglect of such procedure-oriented bias would restrict the debiasing methods to provide a comprehensive solution that is fair in both prediction and procedure.
Therefore, the procedure leading to the decisions is critical for determining whether the decision is fair and should be explicitly considered.
To get a better understanding of this process, we naturally leverage model explainability, which aims to identify the underlying mechanism of why a model makes the prediction and has raised much attention~\cite{CV_explainability}.
The demands on explainability are even more urgent in fairness domain where fairness and explainability are intertwined considering that bias identification largely hinges on the reason of prediction. Additionally, explainability will bring in extra benefit regarding model selection. When two models have comparable performance in utility and traditional fairness, explainability can provide another perspective to select the best model.

To define fairness from an explainability perspective, we first briefly introduce the explanations obtained from explainability methods, denoted as $\mathbf{e}_i$ for the $i$-th instance, which can be the important score per feature (e.g., which features are more salient for the hiring decision)
or other forms. Based on the explanations, an intuitive way for fairness quantification is measuring the distribution difference of $\mathbf{e}_i$ and $\mathbf{e}_j$ where $(i, j) \in (\mathcal{G}_0, \mathcal{G}_1)$. However, the connection between bias and the distribution distance is controversial due to the inherent nature of diverse explanation in real-world and the lack of common understanding on fairness evaluation from attributes where expert knowledge is required and thus limit the automation across domains.
Therefore, we turn the focus from explanation itself to domain-agnostic quality metric.
Explanation quality ($\text{EQ}$) is a measurement of how good the explanation is at interpreting the model and its predictions. Various measurements~\cite{robnik2018perturbation-fidelity} quantify $\text{EQ}$ and provide a broad range of perspectives, which can be applied for calculating fairness. As shown in Figure~\ref{fig.motivation}, the unfairness in the procedure is revealed in EQ. When EQ differs among subgroups, the model treats them discriminatively, providing better-quality explanations for one group than the other. Therefore, EQ gap between subgroups indicates bias and should be mitigated. Inheriting the benefits from the variety of quality metrics, the proposed fairness metric can thus provide different perspectives regarding explainability. 
\subsection{Ratio-based Explanation Fairness ($\Delta_{\text{REF}}$)}

In order to quantify EQ gap, we borrow the idea from traditional fairness metrics $\Delta_{\text{SP}}$\footnote{$\Delta_{\text{SP}}=|P(\hat{y}=1|s=0)-P(\hat{y}=1|s=1)|$}, which quantify bias based on the outputs $\hat{y}_i$. If the proportions of positive predictions are the same for subgroups, the model is fair. Regarding explanation, high quality is a positive label. Similar to the hiring example where each group deserves the same right of being hired, as an opportunity, the positive label should be obtained by subgroups fairly. Therefore, the proportions of instances with high $\text{EQ}$ should be same for fairness consideration. We have the Ratio-based explanation fairness as 
\begin{equation}
    \Delta_{\text{REF}}=|P(\hat{{{q}}}=1|s=0)-P(\hat{{{q}}}=1|s=1)|,
    \label{eq.EFr}
\end{equation}
where $\hat{{{q}}}$ is the quality label for the explanation. If $\text{EQ}$ is high, $\hat{{{q}}}=1$. Otherwise, $\hat{{{q}}}=0$.
However, it would be challenging to define a threshold for high-quality (i.e., when $\text{EQ}$ is larger than the threshold, the explanation is of high-quality) and the range of $\text{EQ}$ might differ across explainers, limiting the generalizability. We transform the threshold to a top $K$ criterion where top $K$ percent of instances with highest EQ are regarded as of high quality (i.e., $\hat{q}_i$ is $1$ when $\text{EQ}_i$ belongs to the top $K$ and $0$ otherwise). The metric $\Delta_{\text{REF}}$ measures the unfairness of high-quality proportion for subgroups and a smaller value relates to a fairer model.

\subsection{Value-based Explanation Fairness ($\Delta_{\text{VEF}}$)}

Under certain cases, bias still exist with a low $\Delta_{\text{REF}}$ due to the ignorance of detailed quality values in the top $K$ (e.g., $\Delta_{\text{REF}}=0$ in Figure~\ref{fig.motivation} where instances in subgroups with high-quality explanations take up the same proportion but instances in $\mathcal{G}_0$ in the top $K$ always have larger $\text{EQ}$ than $\mathcal{G}_1$). 
Therefore, we further compute the difference of  subgroup's average $\text{EQ}$ as Value-based explanation fairness:
\begin{equation}
\Delta_{\text{VEF}}=\bigg|\frac{1}{|\mathcal{G}_0^K|}\sum_{i\in \mathcal{G}_0^K}{\text{EQ}_i}-\frac{1}{|\mathcal{G}_1^K|}\sum_{i\in \mathcal{G}_1^K}{\text{EQ}_i}\bigg|,
    \label{eq.EFv}
\end{equation}
where $\mathcal{G}_s^K$ denotes the top $K$ instances with highest $\text{EQ}$ whose sensitive feature equals $s$. The metric $\Delta_{\text{VEF}}$ measures the unfairness of the average EQ of high-quality instances. Similarly, a smaller value indicates a higher level of fairness.

\subsection{Extending to Multi-class Sensitive Feature}
The previous definitions are formulated for
binary scenario, but they can be easily extended into multi-class sensitive feature scenarios following previous work in result-oriented fairness~\cite{fairwalk,fairdrop}. The main idea is to guarantee the fairness of $\text{EQ}$ across all pairs of sensitive subgroups. Quantification strategies can be applied by leveraging either the variance of $\text{EQ}$ across different sensitive subgroups~\cite{fairwalk} or the maximum difference of $\text{EQ}$ among all subgroup pairs~\cite{fairdrop}.

\section{Comprehensive Fairness Algorithm}\label{sec-alg}
\begin{figure}[t]
     \centering
\includegraphics[width=0.48\textwidth]{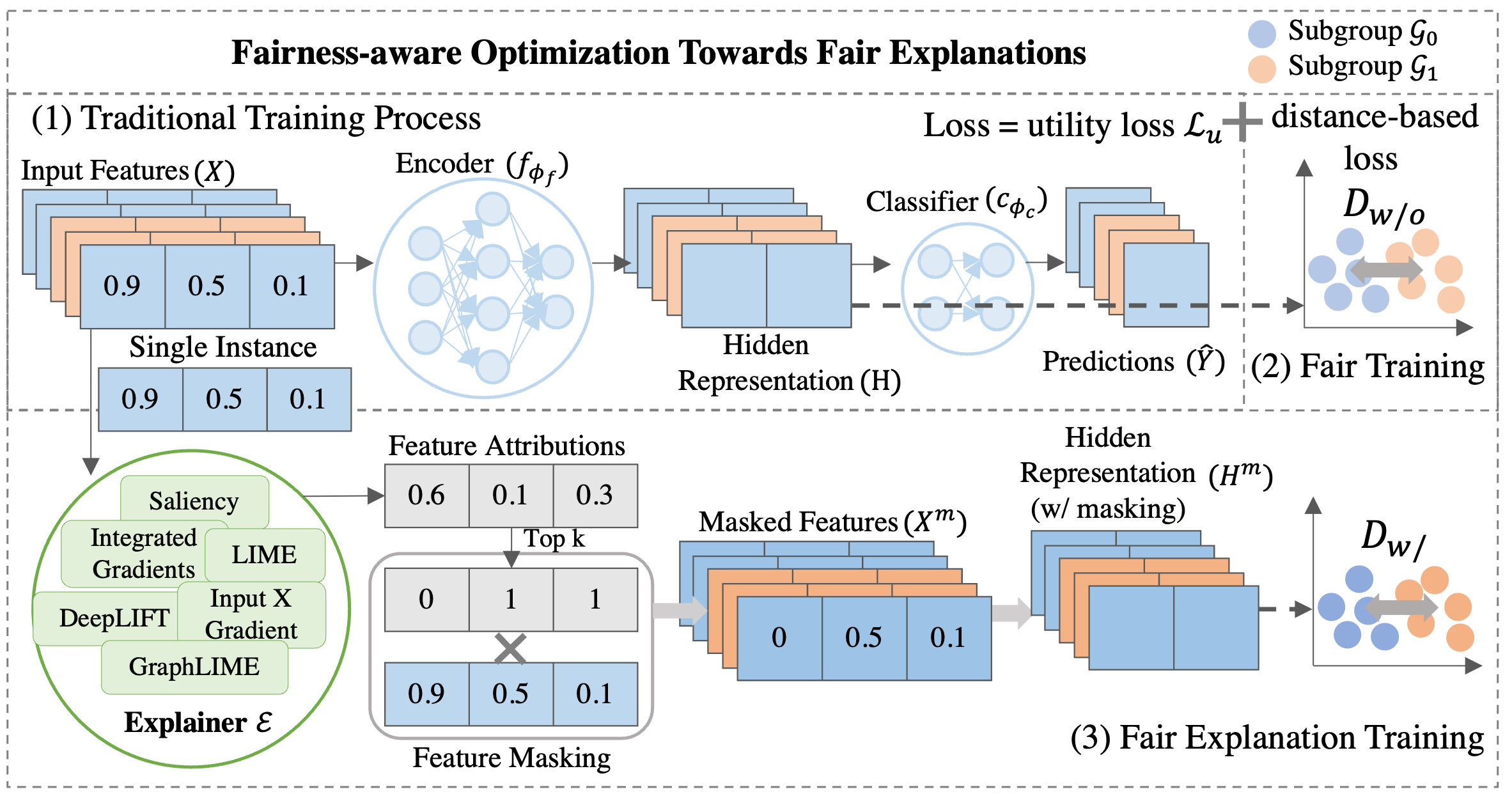}
\vskip -1.5ex
    \caption{Overall framework of CFA.}
    \vskip -2ex
     \label{fig.framework}
\end{figure}

In light of the novel procedure-oriented fairness metrics defined in Section 3, we seek to develop a Comprehensive Fairness Algorithm (CFA) with multiple objectives, aiming to achieve better trade-off performance across not only utility and traditional (result-oriented) fairness, but incorporating our proposed explanation (procedure-oriented) fairness. We note that the explanation fairness is complimentary to the existing traditional fairness metrics as compared to substitutional, and furthermore they all need to be considered under the context of utility performance (as it would be meaningless if having good fairness metrics but unusable utility performance). More specifically, we design the loss for CFA as
\begin{equation}
    \mathcal{L}(\mathcal{G}) = \mathcal{L}_{u}(\mathcal{G}) + \alpha \mathcal{L}_{f}(\mathcal{G}_0, \mathcal{G}_1) +\beta \mathcal{L}_{exp}(\mathcal{G}_0, \mathcal{G}_1),
    \label{overall_loss}
\end{equation}
where $\alpha$, $\beta$ are coefficients to balance these goals, $\mathcal{L}_{u}$ is the utility loss for better utility performance, $\mathcal{L}_f$ and $\mathcal{L}_{exp}$ 
are the traditional and explanation fairness loss designed to mitigate the result/procedure-oriented bias. The framework is shown in Figure~\ref{fig.framework}. In the following part, we first introduce $\mathcal{L}_f$ and $\mathcal{L}_{exp}$, and provide CFA in the end where we find that mitigating procedure-oriented bias improves the result-oriented fairness and thus two loss terms are reduced to one.
\subsection{Traditional Fairness (Result-oriented)}
\label{sec.traditional_opt}
Traditional fairness 
measures whether the output/prediction of the model is biased towards specific subgroups. The prediction is usually classified from the learned representations. Therefore, bias
highly likely originates from the representation space. If the instance representations in subgroups in each class are close, the classifier will make fair decisions. We design an objective function based on the distance:
\begin{equation}
    \mathcal{L}_f(\mathcal{G}_0, \mathcal{G}_1) = \mathbb{E}_{y \in Y}\mathbb{E}_{(i, j)\in (\mathcal{G}_0, \mathcal{G}_1)} \mathcal{D}(\mathbf{h}_i^y, \mathbf{h}_j^y),
    \label{eq.fairness_loss}
\end{equation}
where $y\in Y$ is the utility label, $\mathcal{D}(\cdot, \cdot)$ is a distance metric, $\mathbf{h}_i^y$ and $\mathbf{h}_j^y$ are the hidden representations of instances from subgroups whose utility label equals $y$. By minimizing $\mathcal{L}_f$, instances with same utility label are encouraged to stay close regardless of sensitive attributes.
In Section~\ref{sec.ablation}, we conduct the ablation study to compare the effectiveness of different loss functions, including Sliced-Wasserstein distance (SW), Negative cosine similarity (Cosine), Kullback–Leibler divergence (KL) and Mean square error (MSE). Details about these functions are provided in Appendix~\ref{sec-distance_functions}.

\subsection{Explanation Fairness (Procedure-oriented)}
Explanation fairness measures explanation quality gap between subgroups. To reduce the gap, we first introduce the quality metric used in this paper called fidelity and then introduce the corresponding objective function which aims to minimize the distances between hidden representations of subgroups that are based on features with/without masking.
\subsubsection{Fidelity as Explanation Quality}
Fidelity~\cite{robnik2018perturbation-fidelity} studies the change of prediction performance (e.g., accuracy) after removing the top important features. Intuitively, if a feature is important for the prediction, when removing it, model performance will change significantly. A larger prediction change thus indicates higher importance. Fidelity has two versions - probability/accuracy-based. The former is defined as 
\begin{equation}
    \Delta_{\text{F}_i}= P_\theta(\mathbf{x_i})-P_\theta(\mathbf{x_i}^{\mathbf{m_i}}),
    \label{eq. fidelity}
\end{equation}
where $P_\theta$ is a general model parameterized by $\theta$ aiming to predict the utility label $y_i$, $P_\theta(\mathbf{x_i})$ is the predicted probability of $y_i$, $\mathbf{x_i^m}=\mathbf{x_i} \odot \mathbf{m_i}$ is the masked feature where mask $\mathbf{m_i} = [m_1, m_2, ..., m_d]\in \{0, 1\}^d$ is generated as follows: 
$m_i$ is $0$ when the $i$-th feature has high (i.e., top $k$) important score calculated from explainer $\mathcal{E}$ otherwise $1$. The latter substitutes the probability with binary score indicating whether the predicted label is true. In this paper, we mask the most important feature (i.e., $k=1$) and adopt accuracy-based fidelity for stability consideration. The utility predictor $P_\theta$ is composed of an encoder $f_{\theta_{f}}$ and a classifier $c_{\theta_{c}}$.

\subsubsection{Distance-based Optimization}
When substituting 
explanation quality to fidelity, $\Delta_{\text{VEF}}$ (similar for $\Delta_{\text{REF}}$) becomes
\begin{equation}
\Delta_{\text{VEF}}=\bigg|\frac{1}{|\mathcal{G}_0^K|}\sum_{i\in \mathcal{G}_0^K}{\Delta_{\text{F}_i}}-\frac{1}{|\mathcal{G}_1^K|}\sum_{i\in \mathcal{G}_1^K}{\Delta_{\text{F}_i}}\bigg|.
\label{eq.EFv_F}
\end{equation}

As shown in Eq.~\eqref{eq. fidelity}, each $\Delta_{\text{F}_i}$ consists of two parts which are based on the original feature and the masked feature. We apply the same idea for optimizing traditional fairness in Section~\ref{sec.traditional_opt}. If the hidden representations for instances with/without masking in subgroups are close to each other when utility label is the same, $\Delta_{\text{VEF}}$ will be low. Therefore, to minimize the distance with and without masking, we have the objective function as 
\begin{equation}
\small 
    \mathcal{L}_{exp}(\mathcal{G}_0, \mathcal{G}_1) = \mathbb{E}_{y \in Y}\mathbb{E}_{(i, j)\in (\mathcal{G}_0, \mathcal{G}_1)} (\mathcal{D}(\mathbf{h}_i^y, \mathbf{h}_j^y) + \mathcal{D}(\mathbf{h^m}_i^y, \mathbf{h^m}_j^y)),
    \label{eq.fairness_loss2}
\end{equation}
where $\mathbf{h^m}_i^y$ and $\mathbf{h^m}_j^y$ are the hidden representations based on masked features
with other notations sharing the same meaning in Eq.~\eqref{eq.fairness_loss}. By minimizing $\mathcal{L}_{exp}$, instances with same utility label are encouraged to stay close both before and after masking regardless of their sensitive attributes. 
\subsection{Overall Objective of CFA}
\label{sec.overall_alg}
The above objectives show that $\mathcal{L}_{exp}$ contains $\mathcal{L}_{f}$, which indicates that alleviating procedure-oriented bias helps mitigating result-oriented bias under certain objective design.
The final objective thus becomes $
    \mathcal{L}(\mathcal{G}) = \mathcal{L}_{u}(\mathcal{G}) + \lambda \mathcal{L}_{exp}(\mathcal{G}_0, \mathcal{G}_1)$,
where $\mathcal{L}_{exp}$ aims to mitigate the bias in both predictions and the procedure, and coefficient $\lambda$ flexibly adjusts the optimization towards the desired goal.
Our CFA framework is holistically presented in Algorithm~\ref{alg-general} with its detailed description in Appendix~\ref{sec-alg_in_appendix}.
\SetAlgoNoEnd
\begin{algorithm}[tb]
 \DontPrintSemicolon
 \footnotesize
 \KwIn{Features $\mathbf{X}$ with utility labels $\mathbf{Y}$, top feature proportion $k$, encoder $f_{\theta_{f}}$, classifier $c_{\theta_{c}}$, explainer $\mathcal{E}$, coefficient $\lambda$, learning rate $\epsilon$, distance metric $\mathcal{D}$.}
 
 \While{not converged}{
   
    Hidden representation $\mathbf{H}=f_{\theta_f}(\mathbf{X})$
    
    Prediction $\hat{\mathbf{Y}}=c_{\theta_c}(\mathbf{H})$
    
    Mask $\mathbf{M}=\mathcal{E}(\mathbf{X}, \hat{\mathbf{Y}}, f_{\theta_f}, c_{\theta_c}, $k$)$
    
    Masked feature $\mathbf{X^m}= \mathbf{X} \odot \mathbf{M}$
    
    Hidden representation w/ masking $\mathbf{H}^m=f_{\theta_f}(\mathbf{X}^m)$
    
    $\mathcal{L}_{u}=-\sum_{i=1}^{|\mathbf{Y}|}(y_ilog(\hat{y_i})+(1-y_i)log(1-\hat{y_j}))$

    $D_{w/o} = \mathcal{D} (\mathbf{H}_{\mathcal{G}_0}, \mathbf{H}_{\mathcal{G}_1})$
    
    $D_{w/} = \mathcal{D} (\mathbf{H^m}_{\mathcal{G}_0}, \mathbf{H^m}_{\mathcal{G}_1})$
    
    $\mathcal{L}_{exp} = D_{w/o} + D_{w/}$
  
    $\mathcal{L}=\mathcal{L}_{u}+\lambda \mathcal{L}_{exp}$
    
    $\theta=\theta-\epsilon\nabla_\theta \mathcal{L}$
    
}
\KwRet{A fair utility predictor composed of $f_{\theta_{f}}$ and $c_{\theta_{c}}$}

\caption{\small Comprehensive Fairness Algorithm (CFA)}
\label{alg-general}
\end{algorithm}
\section{Experiments}\label{sec-experiment}
In this section, we conduct extensive experiments to validate the proposed explanation fairness measurements and the effectiveness of the designed algorithm CFA\footnote{Source code available at \url{https://github.com/YuyingZhao/FairExplanations-CFA}}. In particular, we answer the following research questions:
\begin{itemize}
\item \textbf{RQ1}: How well can CFA mitigate the bias related to traditional result-oriented and procedure-oriented explanation fairness measurements (Section~\ref{sec.main_result}, Section~\ref{sec.lambda_effect})?
\item \textbf{RQ2}: How well can CFA balance different categories of objectives compared with other baselines (Section~\ref{sec.tradeoff})?
\item \textbf{RQ3}: Are the fairness metrics and algorithm general to other complex data domains (Section~\ref{sec. graph})?
\end{itemize}
Additionally, we further probe CFA with an ablation study of different loss functions (Section~\ref{sec.ablation}).

\subsection{Experimental Settings}
\subsubsection{Dataset}
We validate the proposed approach on four real-world benchmark datasets: German~\cite{german_dataset}, Recidivism~\cite{bail_dataset}, Math and Por~\cite{math_dataset}, which are commonly adopted for fair ML~\cite{dataset_survey}. Their statistics and the detailed descriptions are given in Appendix~\ref{sec-dataset}.

\subsubsection{Baselines}
Given no prior work has explicitly considered explanation fairness, we compare our methods with MLP model and two fair ML models, namely Reduction~\cite{reductions} and Reweight~\cite{reweight}. To validate the generalizability of our novel defined fairness measures on complex data, we include two fair models in the graph domain called  NIFTY~\cite{2021nifty} and FairVGNN~\cite{fairvgnn}. The details about these methods are provided in Appendix~\ref{sec-baselines}.

\subsubsection{Evaluation Criteria}
We evaluate model performance from four perspectives: model utility, result-oriented fairness,  procedure-oriented explanation fairness, and the overall score. More specifically, the metrics are (1) \textit{Model utility}: area under receiver operating characteristic curve (AUC), F1 score, and accuracy;
(2) \textit{Result-oriented fairness}: two widely-adopted fairness measurements $\Delta_{\text{SP}}$ and $\Delta_{\text{EO}}$\footnote{ $\Delta_{\text{EO}}=|P(\hat{y}=1|y=1,s=0)-P(\hat{y}=1|y=1,s=1)|$}; (3) \textit{Procedure-oriented explanation fairness}: two proposed group fairness metrics $\Delta_{\text{REF}}$ in Eq.~(\ref{eq.EFr}) and $\Delta_{\text{VEF}}$ in Eq.~(\ref{eq.EFv}); (4) \textit{Overall score}: the overall score regarding the previous three measurements via their average (per category): 
\begin{equation}
\scriptsize 
\text{Score}=\frac{\text{AUC}+\text{F1}+\text{Acc}}{3.0}-\frac{\Delta_{\text{SP}}+\Delta_{\text{EO}}}{2.0}-\frac{\Delta_{\text{REF}}+\Delta_{\text{VEF}}}{2.0}.
\end{equation}

\subsubsection{Setup}
For a fair comparison, we record the best model hyperparameters based on the overall score in the validation set. After obtaining the optimal hyperparameters, we then run the model five times on the test dataset to get the average result. The explainer used for all methods is GraphLime~\cite{graphlime} where for general datasets the closest instances are treated as neighbors. The details about hyperparameters are provided in Appendix~\ref{sec-hyper_parameters}.

\vspace{-0.75ex}
\subsection{Performance Comparison}
\label{sec.main_result}
To answer \textbf{RQ1}, we evaluate our proposed method and general fairness methods on real-world datasets and report the utility and fairness measurements along with their standard deviation. The results are presented in Table~\ref{tab-main} where the distance metric $\mathcal{D}$
is Sliced-Wasserstein distance and the ablation study for distance function is in Section~\ref{sec.ablation}. Reduction and Reweight learn a logistic regression classifier to solve the optimization problem, which is different from the MLP-based implementation. They are more stable and have a smaller standard deviation. From the table, we observe:
\begin{itemize}[leftmargin=*]
    \item From the perspective of utility performance, CFA and other fairness methods have comparable performance with the vanilla MLP model, which indicates that little compensation has been made when pursuing higher fairness performance. In some dataset such as Recidivism, the utility performance of CFA is even higher indicating that the distance-based loss might provide a better-quality representation for the downstream tasks.
    \item From the perspective of traditional result-oriented fairness measurements (i.e, $\Delta_{\text{SP}}$ and $\Delta_{\text{EO}}$), CFA has comparable or better performance when compared with existing fairness methods. This shows our model is able to provide fair predictions for subgroups.
    \item From the perspective of procedure-oriented fairness measurements (i.e, $\Delta_{\text{REF}}$ and $\Delta_{\text{VEF}}$), we observe that CFA does not obtain the best performance in all datasets. This is partially due to the complex impact of multiple objectives and the criterion for model selection. When the difference in utility impacts more on the validation score for model selection, then models with higher utility might be selected at a cost of worse explanation fairness.
    \item From the global view, CFA obtains the highest overall score, which measures the multi-task performance, showing its strong ability of balancing multiple goals.
\end{itemize}

\begin{table}[t!]
\scriptsize
\setlength\tabcolsep{1.5pt}
\caption{Model utility and fairness measurements of binary classification. 
The best and second best results are marked \textbf{bold} or \underline{underline}, respectively. Additionally, the $\uparrow$ represents the larger the better and $\downarrow$ represents the opposite.}
\begin{tabular}{c|lcccc}
\Xhline{3\arrayrulewidth}
Dataset & Metric & MLP & Reduction
& Reweight
& CFA \\
\Xhline{3\arrayrulewidth}
\multirow{7}{*}{\rotatebox{90}{\textbf{Recidivism}}} & AUC$\uparrow$ & $86.12\pm1.91$ & $81.17\pm0.00$ & \textbf{89.24} $\boldsymbol{\pm}$ \textbf{0.00} & \underline{$89.02\pm0.86$} \\
& F1$\uparrow$ & $76.54\pm2.52$ & \underline{$76.69\pm0.00$} & $72.99\pm0.00$ & \textbf{81.28} $\boldsymbol{\pm}$ \textbf{1.35} \\
 & Acc$\uparrow$ & {$83.48\pm1.53$} & \underline{$84.66\pm0.00$} & $83.70\pm0.00$ & \textbf{87.17} $\boldsymbol{\pm}$ \textbf{0.84} \\
 \cline{2-6}
 & $\Delta_{\text{SP}}$ $\downarrow$ & {$6.07\pm2.18$} & \underline{{$2.04\pm0.00$}} & $4.27\pm0.00$ & \textbf{1.16} $\boldsymbol{\pm}$ \textbf{0.49} \\
 & $\Delta_{\text{EO}}$ $\downarrow$ & \underline{$3.19\pm0.73$} & $4.66\pm0.00$ & $3.37\pm0.00$ & \textbf{1.14} $\boldsymbol{\pm}$ \textbf{0.39} \\
 \cline{2-6}
 & $\Delta_{\text{REF}}$ $\downarrow$ & $4.45\pm2.96$ & \textbf{0.53} $\boldsymbol{\pm}$ \textbf{0.00} & \underline{$1.34\pm0.91$} & {{$1.98\pm1.23$}} \\
 & $\Delta_{\text{VEF}}$ $\downarrow$ & \underline{$2.1\pm1.38$} & \textbf{2.06} $\boldsymbol{\pm}$ \textbf{0.00} & {$3.22\pm0.00$} & $2.70\pm 0.78$ \\
  \cline{2-6}
& Score $\uparrow$ & $74.15\pm2.03$ & \underline{{$76.19\pm0.00$}} & {$75.88\pm0.00$} & \textbf{82.33} $\boldsymbol{\pm}$ \textbf{0.62} \\
 \Xhline{3\arrayrulewidth}
 
\multirow{7}{*}{\rotatebox{90}{\textbf{German}}} &  AUC$\uparrow$ & \underline{{$66.77\pm2.07$}} & {$63.95\pm0.05$} & \textbf{68.03} $\boldsymbol{\pm}$ \textbf{0.00} & $60.92\pm5.18$ \\
& F1$\uparrow$ & $71.11\pm3.48$ & $72.48\pm1.15$ & \underline{{$74.71\pm0.00$}} & \textbf{81.14} $\boldsymbol{\pm}$ \textbf{2.29}\\
 & Acc$\uparrow$ & {$63.28\pm3.23$} & $64.40\pm1.41$ & \underline{{$65.60\pm0.00$}} & \textbf{70.00} $\boldsymbol{\pm}$ \textbf{2.96} \\
 \cline{2-6}
 & $\Delta_{\text{SP}}$ $\downarrow$ & $39.80\pm9.32$ & $24.55\pm1.76$ & \underline{$20.53\pm0.00$} & \textbf{7.21} $\boldsymbol{\pm}$ \textbf{6.42} \\
 & $\Delta_{\text{EO}}$ $\downarrow$ & $31.39\pm11.49$ & $16.03\pm3.86$ & \underline{{$12.18\pm0.00$}} & \textbf{4.60} $\boldsymbol{\pm}$ \textbf{4.08} \\
 \cline{2-6}
 & $\Delta_{\text{REF}}$ $\downarrow$ & \textbf{4.50} $\boldsymbol{\pm}$ \textbf{2.81} & $16.09\pm8.59$ & \underline{{$7.16\pm0.00$}} & {{$10.02\pm4.24$}} \\
 & $\Delta_{\text{VEF}}$ $\downarrow$ & \textbf{8.85} $\boldsymbol{\pm}$ \textbf{8.98} & $23.55\pm7.67$ & {$16.67\pm0.00$} & \underline{$12.87\pm9.04$} \\
   \cline{2-6}
& Score $\uparrow$ & $24.78\pm12.33$ & {$26.83\pm4.09$} & \underline{{$41.18\pm0.00$}} & \textbf{53.34} $\boldsymbol{\pm}$ \textbf{7.27} \\
 \Xhline{3\arrayrulewidth}
 
\multirow{7}{*}{\rotatebox{90}{\textbf{Por}}} &  AUC$\uparrow$ & \underline{{$90.86\pm0.35$}} & {$67.64\pm0.00$} & $89.07\pm0.00$ & \textbf{91.30} $\boldsymbol{\pm}$ \textbf{0.55} \\
& F1$\uparrow$ & {\underline{$58.41\pm4.10$}} & {$51.43\pm0.00$} & $51.43\pm 0.00$ & \textbf{60.55} $\boldsymbol{\pm}$ \textbf{4.73} \\
 & Acc$\uparrow$ & {\underline{$89.57\pm0.78$}} & \underline{{$89.57\pm0.00$}} & \underline{$89.57\pm0.00$} & \textbf{89.82} $\boldsymbol{\pm}$ \textbf{1.00} \\
 \cline{2-6}
 & $\Delta_{\text{SP}}$ $\downarrow$ & $2.08\pm0.75$ & \underline{{$1.93\pm0.00$}} & \underline{$1.93\pm0.00$} & \textbf{1.00} $\boldsymbol{\pm}$ \textbf{0.72} \\
 & $\Delta_{\text{EO}}$ $\downarrow$ & $32.35\pm7.07$ & \textbf{20.59} $\boldsymbol{\pm}$ \textbf{0.00} & \textbf{20.59} $\boldsymbol{\pm}$ \textbf{0.00} & \underline{{$27.65\pm 5.44$}} \\
 \cline{2-6}
 & $\Delta_{\text{REF}}$ $\downarrow$ & {$8.68\pm3.18$} & \textbf{1.37} $\boldsymbol{\pm}$ \textbf{0.00} & $8.68\pm0.00$ & \underline{{$4.66\pm3.76$}} \\
 & $\Delta_{\text{VEF}}$ $\downarrow$ & \underline{$4.44\pm2.22$} & \textbf{0.00} $\boldsymbol{\pm}$ \textbf{0.00} & $7.69\pm0.00$ & {$4.70\pm3.67$}\\
  \cline{2-6}
 & Score $\uparrow$ & $55.83\pm3.97$ & \underline{{$57.60\pm0.00$}} & {$57.25\pm0.00$} & \textbf{61.55} $\boldsymbol{\pm}$ \textbf{3.26} \\
 \Xhline{3\arrayrulewidth}
 
 \multirow{7}{*}{\rotatebox{90}{\textbf{Math}}} & AUC$\uparrow$ & \underline{{$95.73\pm1.82$}} & {$86.44\pm0.06$} & $95.06\pm1.06$ & \textbf{96.97} $\boldsymbol{\pm}$ \textbf{0.55} \\
 & F1$\uparrow$ & \underline{{$83.32\pm3.32$}} & {$82.65\pm0.21$} & $82.17\pm2.70$ & \textbf{86.74} $\boldsymbol{\pm}$ \textbf{1.74} \\
 & Acc$\uparrow$ & {$88.60\pm2.80$} & \underline{{$89.00\pm0.00$}} & $88.00\pm2.45$ & \textbf{91.00} $\boldsymbol{\pm}$ \textbf{1.26} \\
 \cline{2-6}
 & $\Delta_{\text{SP}}$ $\downarrow$ & \textbf{2.14} $\boldsymbol{\pm}$ \textbf{1.64} & \underline{$3.59\pm1.57$} & $4.54\pm1.59$ & {$4.59\pm 2.40$} \\
 & $\Delta_{\text{EO}}$ $\downarrow$ & \underline{$16.89\pm4.73$} & {$28.00\pm2.67$} & $23.78\pm3.27$ & \textbf{6.22} $\boldsymbol{\pm}$ \textbf{3.19} \\
 \cline{2-6}
 & $\Delta_{\text{REF}}$ $\downarrow$ & $9.60\pm5.31$ & \textbf{2.08} $\boldsymbol{\pm}$ \textbf{2.56} & \underline{$4.8\pm0.00$} & {$6.08\pm5.56$} \\
 & $\Delta_{\text{VEF}}$ $\downarrow$ & {$4.22\pm5.18$} & {$8.22\pm4.13$} & \textbf{3.56} $\boldsymbol{\pm}$ \textbf{4.35} & \underline{$4.00\pm 4.90$} \\
   \cline{2-6}
 & Score $\uparrow$ & \underline{$72.79\pm3.76$} & {$65.08\pm1.50$} & {$70.08\pm6.67$} & \textbf{81.12} $\boldsymbol{\pm}$ \textbf{5.59} \\
 \Xhline{3\arrayrulewidth}
\end{tabular}
\vskip -2ex
\label{tab-main}
\end{table}

\vspace{-0.75ex}
\subsection{Tradeoff between Different Objectives}
\label{sec.tradeoff}
\begin{figure}[h!]
     \centering
\includegraphics[width=0.45\textwidth]{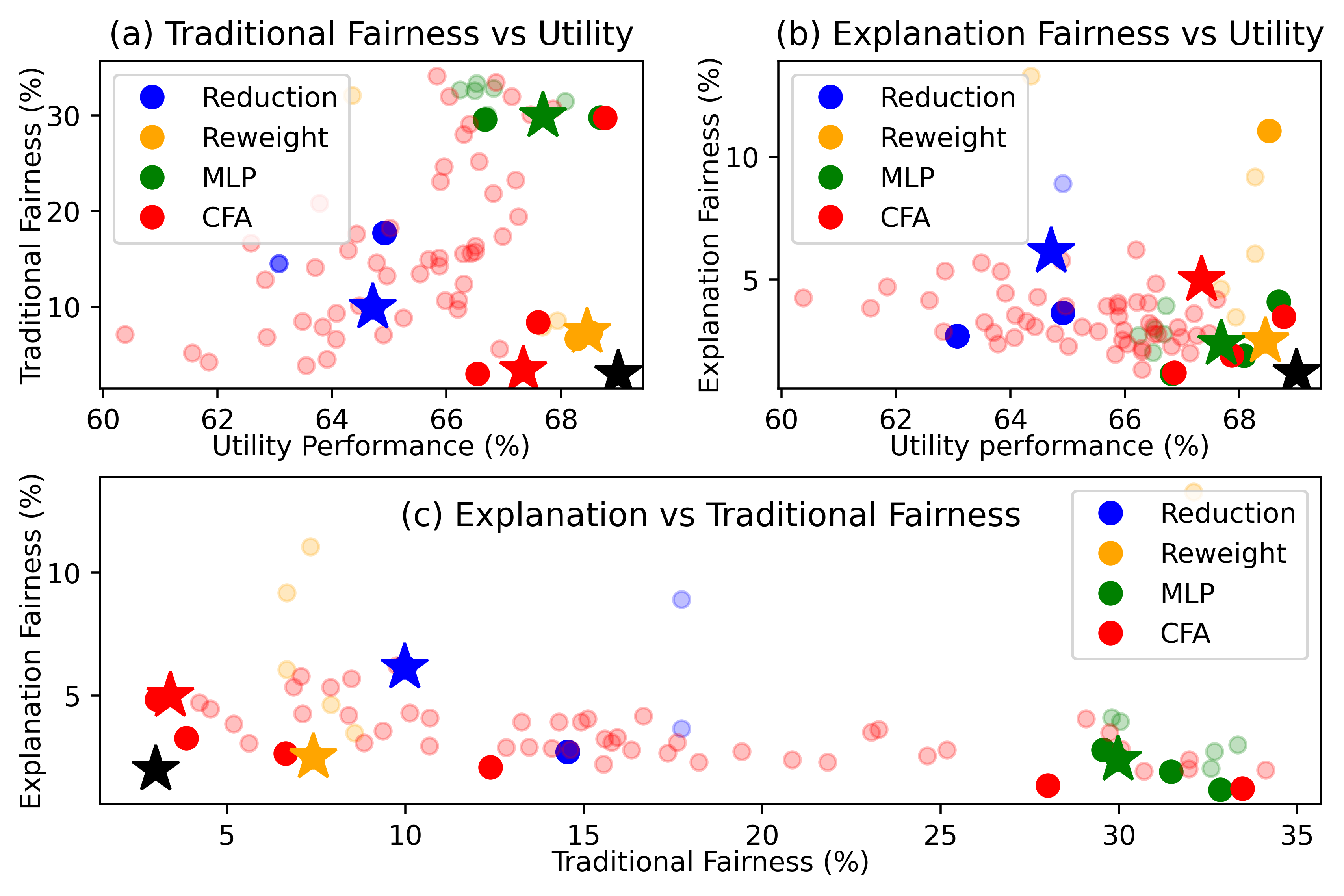}
\vskip -1.5ex
     \caption{Performance comparison in German dataset.}
     \label{fig.tradeoff}
     \vskip -2ex
\end{figure}

To answer \textbf{RQ2}, we explore the tradeoff between objectives in the validation records. Figure~\ref{fig.tradeoff} shows the performance comparison in German dataset. Points relate to specific hyperparameters. Larger and darker circles are non-dominated solutions which form the Pareto frontier in multi-task and the others are dominated which either have poorer accuracy or fairness. Black star indicates the optimal solution direction and other stars correspond to the best hyperparameter setting based on the overall score. For Reduction, only three different results are obtained, showing its inability to provide
diverse solutions. MLP model has poor performance in traditional fairness and achieve varing levels of explanation fairness. But if not selected based on the overall score, the model will gain slight improvement in utility performance at the cost of fairness, which validates the benefit of using overall score for model selection. While Reweight provides good performance in traditional fairness, most of its settings are unable to provide predictions with good explanation fairness, indicating that alleviating result-oriented bias does not necessarily mitigate procedure-oriented bias and this type of bias will be ignored by traditional fairness. This further highlights the necessity of considering explainability to provide a procedure-oriented perspective. CFA provides diverse solutions and thus satisfies different needs for utility and fairness. Although not always the one closest to the optimal solution, it is the only one method that maintains close to the optimal in all three conditions, indicating CFA has the best performance regarding balancing difference objectives which is also revealed in the overall score in Table~\ref{tab-main}.

\begin{figure}[t]
     \centering
\includegraphics[width=0.49\textwidth]{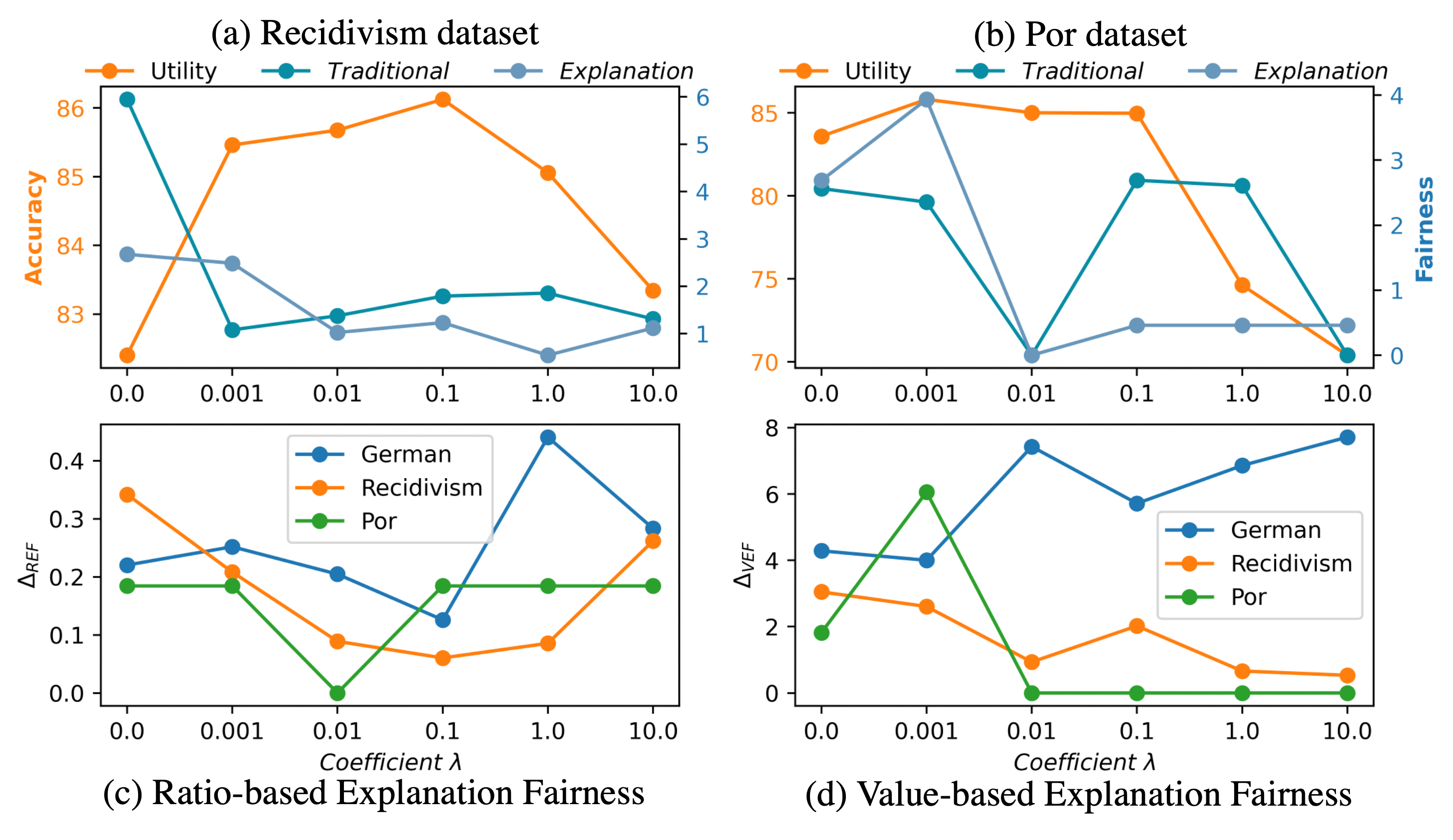}
\vskip -1.25ex
     \caption{The effect of the coefficient $\lambda$. }
     \label{fig.lambda}
     \vskip -2.25ex
\end{figure}

\subsection{Effect of The Coefficient $\lambda$}
\label{sec.lambda_effect}
We explore the impact of the regularization coefficient $\lambda$ which is used for balancing utility and fairness performance. Results based on the validation records are in Figure~\ref{fig.lambda} where (a)-(b) record the average performance of utility, traditional/explanation fairness and (c)-(d) present the detailed explanation fairness $\Delta_{\text{REF}}$ and $\Delta_{\text{VEF}}$.
Figure~\ref{fig.lambda} (b) shows that fairness performance increases while utility performance decreases along with the increase of $\lambda$ in Por dataset.
While in Figure~\ref{fig.lambda} (a), fairness performance improves but utility performance also increases at first in Recidivism dataset, which indicates that in some scenario, good utility and fairness performance can exist together. 
From Figure~\ref{fig.lambda} (c)-(d),
we observe that explanation fairness decreases for Recidivism dataset. 
The trend is not clear for other datasets but we can find at least one point with better performance
than that without fairness optimization which verifies that 
the fairness objective is beneficial for improving explanation fairness.

\subsection{Graph Domain Extension}
\label{sec. graph}
\begin{figure}[t]
     \centering
\includegraphics[width=0.47\textwidth]{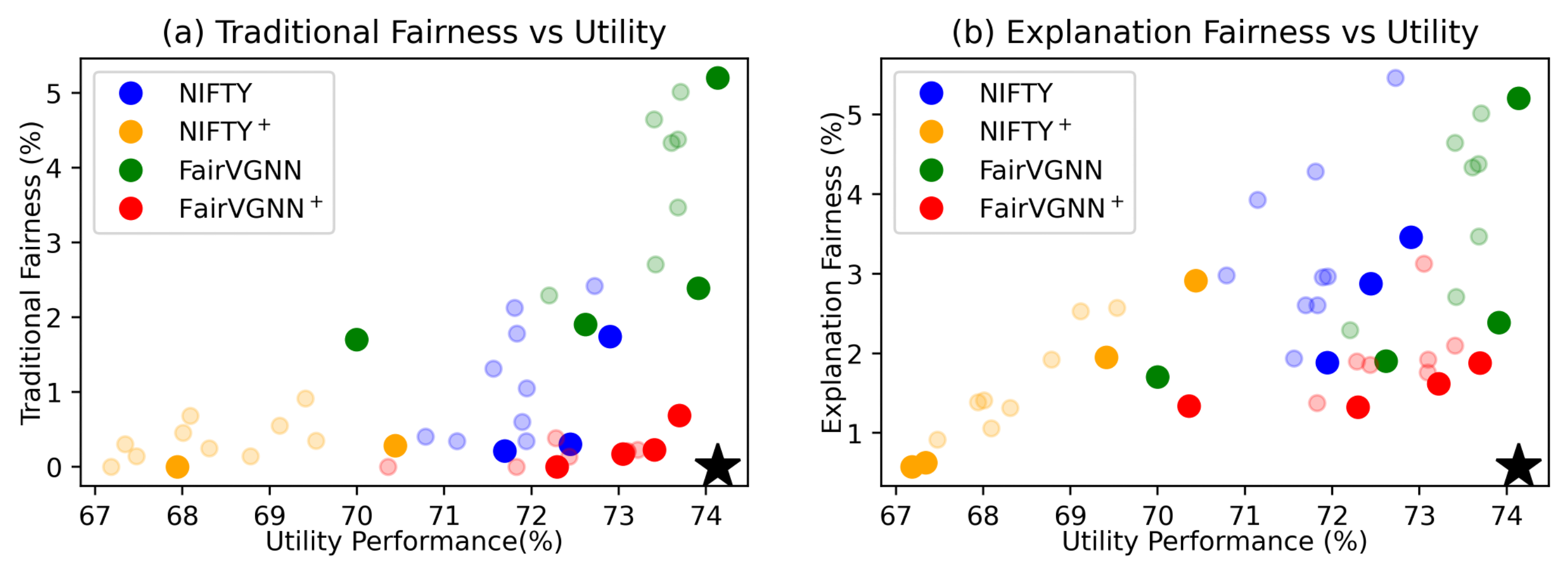}
\vskip -1.25ex
    \caption{Performance comparison between graph-based methods with and without fair explanation optimization.}
     \label{fig.graph}
     \vskip -0.75ex
\end{figure}

To answer \textbf{RQ3}, we take graph as an example to show that our metrics are universal
and can be applied to various domains. To handle the issue of feature propagation which results in sensitive information leakage, masking strategy is slightly different compared with i.i.d. data where mask is applied per instance. In graph domain, the corresponding channels in the neighborhood are also masked. We obtained results for binary classification for NIFTY~\cite{2021nifty} and FairVGNN~\cite{fairvgnn}. Figure~\ref{fig.graph} (b) shows these methods with fair consideration have explanation bias encoded in the procedure. 

Additionally, to test the effectiveness of proposed optimization, fairness loss term in Eq.~\eqref{eq.fairness_loss} is added to the objective function (denoted as $\text{NIFTY}^+$ and $\text{FairVGNN}^+$). Figure~\ref{fig.graph} shows performance in traditional/explanation fairness both improve, indicating that optimization is applicable to graph and successfully alleviates bias from both predictions and procedure. The impacts on utility performance are different - NIFTY experiences a larger decrease than FairVGNN due to a lack of coefficient tuning of the newly added term, which leads to the difference in Figure~\ref{fig.graph}.
The result shows the potential of improving explanation fairness and tradeoff can be adjusted later by tuning the coefficient.

\subsection{Ablation Study}
\label{sec.ablation}
We conduct ablation study on the key component of the fairness objective function, using different distance functions. For ease of experiment, we only tune $\lambda$ with other hyperparameters fixed to reduce the search space. Results in Table~\ref{tab-loss} show that all of them obtain best final score when compared with baseline results in Table~\ref{tab-main}, validating that CFA improves fairness despite of the choices of the distance measurements. Regarding different distance metrics, SW achieves better performance in fairness aspects than KL owing to the knowledge of underlying space geometry; MSE has poor utility performance since enforcing groups to have same embeddings is not necessary; while Cosine is hypothesized to have the best performance due to ignoring scale which inherently relaxes the restriction enforced by MSE.

\begin{table}[t]
\scriptsize
\setlength{\extrarowheight}{.5pt}
\setlength\tabcolsep{1.5pt}
\caption{Performance of different distance functions.}
\vskip -1.75ex
\begin{tabular}{c|lcccc}
\Xhline{3\arrayrulewidth}
Dataset & Metric & SW & Cosine & KL & MSE \\
\Xhline{3\arrayrulewidth}
\multirow{7}{*}{\rotatebox{90}{\textbf{Recidivism}}} & AUC$\uparrow$ & $88.35\pm0.49$ & \textbf{91.73} $\boldsymbol\pm$ \textbf{2.82} & \underline{$90.00\pm1.95$} & $89.43\pm1.03$ \\
& F1$\uparrow$ & $79.82\pm0.61$ & \underline{$82.03\pm3.11$} & \textbf{82.26} $\boldsymbol\pm$ \textbf{3.91} & $81.08\pm1.78$ \\
 & Acc$\uparrow$ & {$86.13\pm0.21$} & $87.52\pm2.40$ & \textbf{87.80} $\boldsymbol\pm$ \textbf{2.64} & \underline{$87.55\pm0.99$} \\
 \cline{2-6}
 & $\Delta_{\text{SP}}$ $\downarrow$ & \textbf{0.60} $\boldsymbol\pm$ \textbf{0.43} & {$3.54\pm2.11$} & $3.92\pm2.35$ & \underline{$1.98\pm 1.76$} \\
 & $\Delta_{\text{EO}}$ $\downarrow$ & \textbf{1.18} $\boldsymbol\pm$ \textbf{0.58} & \underline{$1.71\pm1.40$} & $4.65\pm2.74$ & {$4.02\pm1.05$} \\
 \cline{2-6}
 & $\Delta_{\text{REF}}$ $\downarrow$ & \underline{$1.50\pm0.98$} & \textbf{1.35} $\boldsymbol\pm$ \textbf{1.15} & $5.46\pm3.54$ & {$2.31\pm2.23$} \\
 & $\Delta_{\text{VEF}}$ $\downarrow$ & \underline{$1.92\pm1.98$} & \textbf{1.71} $\boldsymbol\pm$ \textbf{1.40} & {$4.65\pm2.74$} & $3.01\pm 2.52$ \\
  \cline{2-6}
& Score $\uparrow$ & \underline{$82.17\pm0.93$} & \textbf{82.71} $\boldsymbol\pm$ \textbf{4.01} & {$78.44\pm4.81$} & $80.36\pm1.61$ \\
 \Xhline{3\arrayrulewidth}
 \end{tabular}
 \vskip -2.5ex
\label{tab-loss}
\end{table}

\section{Conclusion}\label{sec-conclusion}
In this paper, we investigate the potential bias during the decision-making procedure, which is ignored by traditional metrics. We provide a novel fairness perspective to raise the concern of such procedure-oriented bias. We utilize explainability to provide insights into the procedure and identify bias with two novel fairness metrics based on explanation quality. To simultaneously fulfill multiple goals - improving traditional fairness, satisfying explanation fairness, and maintaining utility performance, we design a Comprehensive Fairness Algorithm (CFA) for optimization. During the optimization, we uncover that optimizing procedure-oriented fairness is beneficial for result-oriented fairness. Experimental results demonstrate that our proposed explanation fairness captures bias ignored by previous result-oriented metrics, and the designed CFA effectively mitigates bias from multiple perspectives while maintaining good model utility. Additionally, our proposed metrics and the optimization strategy can be easily applied to other domains, showing a good generalizability. In the future, we plan to explore explanation fairness in inherently explainable models and further design fair and explainable GNNs. 
We also plan to extend and apply CFA towards fair model explanations in other data types (e.g., images and text) and in the setting of explanation supervision~\cite{gao2022res}.

\balance
\bibliography{ms.bib}
\clearpage 
\appendix

\section{Appendix}\label{sec-appendix}

\subsection{Notations}\label{sec-notation}

Table~\ref{tb:symbols} summarizes notations used throughout the paper.

\subsection{Distance Function Details}
\label{sec-distance_functions}
Different distance functions can be applied to define the distance-based loss, here we provide the description for four distance measurements: Sliced-Wasserstein distance (SW)~\cite{SW0,SW1}, Negative cosine similarity (Cosine), Kullback–Leibler divergence (KL) and Mean square error (MSE).

\begin{itemize}
    \item \textbf{Sliced-Wasserstein distance (SW)} measures the distribution distance, which has a similar property with Wasserstein distance but is more efficient in computation. The main idea is to transform high dimensional distributions for representations through linear projections (via Radon transform) to one dimension and calculate the Wasserstein distance on the one dimensional distributions for input subgroups. The distance is defined as: 

\begin{equation}
\small
    \mathcal{D}_{\text{SW}}(\mathcal{G}_0, \mathcal{G}_1) = \mathbb{E}_{y \in Y} \int_{\mathbb{S}^{d-1}}W(\mathcal{R}\mathbf{H}_{\mathcal{G}_0}^y(.,\theta),\mathcal{R}\mathbf{H}_{\mathcal{G}_1}^y(.,\theta))\dif\theta,
    \label{eq.sw}
\end{equation}
where $\mathbb{S}^{d-1}$ is the unit sphere in $\mathbb{R}^d$ where $d$ is the feature dimension, $Y$ is the utility labels, $\mathbf{H}_{\mathcal{G}_0}^y$ and $\mathbf{H}_{\mathcal{G}_1}^y$ are the hidden representations of instances in subgroup $\mathcal{G}_0$ and $\mathcal{G}_1$ with utility label $y$. 
\end{itemize}

For the other distance measurements, we give a unified definition.

Suppose there are $N_s$ points sampled from $\mathcal{G}_0$ and $\mathcal{G}_1$ (i.e., $\mathcal{G}_0^s\subset{\mathcal{G}_0}$, $\mathcal{G}_1^s\subset{\mathcal{G}_1}$, and $|\mathcal{G}_0^s|=|\mathcal{G}_1^s|=N_s$). The hidden representations corresponding to the sampled instances in the subgroups are denoted as $\mathbf{H}_{\mathcal{G}_0^s}$ and $\mathbf{H}_{\mathcal{G}_1^s}$.
\begin{itemize}
\item \textbf{Negative cosine similarity (Cosine)} is the negative of computed cosine similarity, which is defines as:
\begin{equation}
    \mathcal{D}_{\text{Cosine}}=-\frac{1}{N_s}\|\frac{\mathbf{H}_{\mathcal{G}_0^s} \cdot \mathbf{H}_{\mathcal{G}_1^s}}{max(\|\mathbf{H}_{\mathcal{G}_0^s}\|^2, \|\mathbf{H}_{\mathcal{G}_1^s}\|^2, \epsilon)}\|_1
\end{equation}
where $\epsilon$ is a small value to avoid division by zero and $\|\cdot\|_1$ denotes $L^1$-norm.
\item \textbf{Kullback–Leibler divergence (KL)} computes the point-wise divergence as: 
\begin{equation}
    \mathcal{D}_\text{KL}=\frac{1}{N_s}\|\mathbf{H}_{\mathcal{G}_0^s} \cdot log\frac{\mathbf{H}_{\mathcal{G}_0^s}}{\mathbf{H}_{\mathcal{G}_1^s}}\|_1
\end{equation}
\item \textbf{Mean square error (MSE)} measures the node-wise distance, which is defined as:
\begin{equation}
    \mathcal{D}_{\text{MSE}}(\mathcal{G}_0, \mathcal{G}_1) = \frac{1}{N_s}\|\mathbf{H}_{\mathcal{G}_0^s}-\mathbf{H}_{\mathcal{G}_1^s}\|_2
\end{equation}
where $\|\cdot\|_2$ denotes $L^2$-norm.
\end{itemize}

\begin{table}[H]
\small
\setlength\tabcolsep{3pt}
\setlength{\extrarowheight}{.15pt}
\caption{Notations commonly used in this paper and the corresponding descriptions.}
\label{tb:symbols}
\begin{tabular}{p{0.1\textwidth}p{0.38\textwidth}}
\hline
\textbf{Notations}       & \textbf{Definitions or Descriptions} \\
\hline
$\Delta_{\text{REF}}$/$\Delta_{\text{VEF}}$ & Ratio-based/Value-based Explanation Fairness\\
$\Delta_{\text{SP}}$ & Statistical Parity \\
$\Delta_{\text{EO}}$ & Equality of Opportunity\\
\hline
$s$ & Sensitive label $s \in \{0, 1\}$, e.g., female and male \\
$y$/$Y$ & Utility label $y \in Y$\\
$K$ & The hyperparameter to assign top $K$ ratio instances with highest explanation quality as high, otherwise low \\
$k$ & The number of features which will be masked to zero \\
\hline

$\mathcal{G}$ & The whole group \\
$\mathcal{G}_0/\mathcal{G}_1$ & Subgroups based on sensitive label, for example, the whole group is splitted into female and male subgroup based on gender \\
$\mathcal{G}_s^K$ & Subgroups with sensitive label $s$ which has high-quality explanation labelled based on $K$\\
\hline
$\text{EQ}_i$ & The explanation quality of $i$-th instance, $\text{EQ}_i\in{ \left[ 0, 1 \right] }$\\
$\hat{y}_i$ & The predicted utility label for $i$-th instance \\
$\hat{q}_i$ & The explanation quality label for $i$-th instance, e.g., high/low, $\hat{q}_i \in \{0, 1\}$\\
$\mathbf{e}_i$ & The explanation for $i$-th instance\\
$\Delta_{\text{F}_i}$ & The fidelity for $i$-th instance\\
\hline
$d$ & Feature dimension \\
$d^h$ & Dimension of hidden layer \\
$N$ & The number of whole group \\
$N_0$/$N_1$ & The number of subgroups $G_0$ and $G_1$ \\
$\mathbf{h}_i/\mathbf{H}$ & The hidden representation from the last layer (i.e., the representation outputted from the encoder), $\mathbf{h}_i\in \mathbb{R}^{d^h}$, $\mathbf{H}\in \mathbb{R}^{N \times d^h}$ \\
$\mathbf{H}_{G_0}$/$\mathbf{H}_{G_1}$ & The hidden representations stacked with instances in $G_0$/$G_1$, $\mathbf{H}_{G_0}\in \mathbb{R}^{N_0 \times d^h}$, $\mathbf{H}_{G_1}\in \mathbb{R}^{N_1 \times d^h}$ \\
$N_s$ & The number of sample\\
$\hat{\mathbf{H}}_{G_0}$/$\hat{\mathbf{H}}_{G_1}$ & The hidden representations sampled from $\mathbf{H}_{G_0}$/$\mathbf{H}_{G_1}$, $\mathbf{H}_{G_1}^s\in \mathbb{R}^{N_s \times d^h}$ , $\mathbf{H^s}_{G_1}\in \mathbb{R}^{N_s \times d^h}$ \\
$\mathbf{h}_i^y$ & The hidden embedding of $i$-th instance whose utility label equals $y$ 
\\
$\mathbf{m}_i$ & The mask for $i$-th instance, $\mathbf{m}_i \in \{0, 1\}^d$ \\
$\mathbf{x}_i$ & The input representation for $i$-th instance, $\mathbf{x}_i \in \mathbb{R}^d$ \\
$\mathbf{x}_i^m$ & The masked representation for $i$-th instance, $\mathbf{x}^m_i \in \{0, 1\}^d$, $\mathbf{x}_i^m = \mathbf{x}_i \odot \mathbf{m}_i$ \\
\hline
$\mathcal{D}(\cdot, \cdot)$ & Distance metric \\
$D_{w/}$/$D_{w/o}$ & The distance between embeddings based on the representation with/without masking \\
\hline
$\mathcal{L}_u$ & Loss term for improving utility performance \\
$\mathcal{L}_f$ & Loss term for improving traditional fairness \\
$\mathcal{L}_{exp}$ & Loss term for improving explanation fairness \\
$\alpha$/$\beta$/$\lambda$ & Regularization coefficients\\
\hline
$\mathcal{E}$ & The explainer\\
$f_{\theta_f}$ & The encoder which generates the hidden representations \\
$c_{\theta_c}$ & The classifier which makes the prediction based on learned representation from the encoder\\

\end{tabular}
\vskip 15ex
\end{table}

\SetAlgoNoEnd
\begin{algorithm}[tb]
 \DontPrintSemicolon
 \footnotesize
 \KwIn{Features $\mathbf{X}$ with utility labels $\mathbf{Y}$, top feature proportion $k$, encoder $f_{\theta_{f}}$, classifier $c_{\theta_{c}}$, explainer $\mathcal{E}$, coefficient $\lambda$, learning rate $\epsilon$, distance metric $\mathcal{D}$.}
 
 \While{not converged}{
   
    Hidden representation $\mathbf{H}=f_{\theta_f}(\mathbf{X})$
    
    Prediction $\hat{\mathbf{Y}}=c_{\theta_c}(\mathbf{H})$
    
    Mask $\mathbf{M}=\mathcal{E}(\mathbf{X}, \hat{\mathbf{Y}}, f_{\theta_f}, c_{\theta_c}, $k$)$
    
    Masked feature $\mathbf{X^m}= \mathbf{X} \odot \mathbf{M}$
    
    Hidden representation w/ masking $\mathbf{H}^m=f_{\theta_f}(\mathbf{X}^m)$
    
    $\mathcal{L}_{u}=-\sum_{i=1}^{|\mathbf{Y}|}(y_ilog(\hat{y_i})+(1-y_i)log(1-\hat{y_j}))$

    $D_{w/o} = \mathcal{D} (\mathbf{H}_{\mathcal{G}_0}, \mathbf{H}_{\mathcal{G}_1})$
    
    $D_{w/} = \mathcal{D} (\mathbf{H^m}_{\mathcal{G}_0}, \mathbf{H^m}_{\mathcal{G}_1})$
    
    $\mathcal{L}_{exp} = D_{w/o} + D_{w/}$
  
    $\mathcal{L}=\mathcal{L}_{u}+\lambda \mathcal{L}_{exp}$
    
    $\theta=\theta-\epsilon\nabla_\theta \mathcal{L}$
    
}
\KwRet{A fair utility predictor composed of $f_{\theta_{f}}$ and $c_{\theta_{c}}$}

\caption{\small Comprehensive Fairness Algorithm (CFA)}
\label{alg-general-copy}
\end{algorithm}

\begin{algorithm}[htbp]
 \DontPrintSemicolon
 \footnotesize
 \KwIn{Hidden representations $\mathbf{H}_{\mathcal{G}_0}, \mathbf{H}_{\mathcal{G}_1}$, slice number $I$}
    Init $D=0$
    
    \For{i $\leftarrow 1$ \KwTo $I$}
    { 
    Sample random unit vectors $\hat{\mathbf{v}}_l$
    
    $\widetilde{\mathbf{H}}^s_{\mathcal{G}_0}\leftarrow\{\hat{\mathbf{v}}_i^T\mathbf{H}_{\mathcal{G}_0}\}_{n=1}^{N}$,
    
    $\widetilde{\mathbf{H}}^s_{\mathcal{G}_1}\leftarrow\{\hat{\mathbf{v}}_i^T\mathbf{H}_{\mathcal{G}_1}\}_{n=1}^{N}$
    
    $\widetilde{\mathbf{X}}^\pi_{\mathcal{G}_0}\leftarrow sort(\widetilde{\mathbf{H}}^s_{\mathcal{G}_0}), \widetilde{\mathbf{H}}^\pi_{\mathcal{G}_1}\leftarrow sort(\widetilde{\mathbf{H}}^s_{\mathcal{G}_1})$
    
     $D = D+\frac{1}{N}\left\lVert \widetilde{\mathbf{H}}^\pi_{\mathcal{G}_0}-\widetilde{\mathbf{H}}^\pi_{\mathcal{G}_1}\right\rVert^2$
    
    }
    \KwRet{$D$};
\caption{\small Sliced-Wasserstein Distance}
 \label{alg-sw}
\end{algorithm}

\subsection{Algorithm Details}
\label{sec-alg_in_appendix}

The detailed algorithm for CFA is presented in Algorithm~\ref{alg-general-copy} (Note: this is the same 
as Algorithm~\ref{alg-general} in the main text and included here for ease of reading.)
During each training epoch (lines 2 to 14), we obtain the hidden representations (line 2) and prediction (line 3) which are fed into the explainer to obtain the mask (line 4). The masked feature is then generated by the hadamard product of the original feature and the mask (line 5). Based on the masked feature, the hidden representation w/ masking is obtained (line 6). Then, $\mathcal{L}_u$ is calculated on line 7 based on the prediction and ground truth. The representation distances between subgroups without ($D_{w/o}$) and with masking ($D_{w/}$) are computed based on the representations according to the given distance function $\mathcal{D}$ on line 8 to 9. Finally, after obtaining each term, they are aggregated and back-propagated to optimize the parameters in the encoder and the classifier (line 11 to 12).

Specifically, for Sliced-Wasserstein Distance, we follow Algorithm~\ref{alg-sw} where we approximate the integral in Eq.~\eqref{eq.sw} with a simple Monte Carlo scheme that draws samples from
the uniform distribution and replace the integral with a finite summation~\cite{kolouri2019generalized}.

\subsection{Dataset Details}
\label{sec-dataset}
The datasets used in this paper are as follows:
\begin{itemize}
    \item \textbf{German}~\cite{german_dataset}: The task is to classify the credit risk of the clients in a German bank as high or low, with gender being the sensitive feature.
    \item \textbf{Recidivism}~\cite{bail_dataset}: The task is to predict whether a defendant would be more likely to commit a crime once released on bail, with race as the sensitive feature.
    
    \item \textbf{Student-Portuguese (Por)} / \textbf{Mathematics (Math)} \cite{math_dataset}: The task is to predict whether a student's final year grade of course Math/Portuguese is high, with gender as the sensitive feature.
\end{itemize}
Their statistics are shown in Table~\ref{tab-dataset} where Math and Por datasets are the same students but differ in their class labels, i.e., performance in a Math or Portuguese course.

\begin{table}[tbp]
\footnotesize
\setlength{\extrarowheight}{.095pt}
\setlength\tabcolsep{3pt}
\caption{Dataset statistics.}
\vskip -1ex
\centering
\begin{tabular}{lccccc}
 \Xhline{2\arrayrulewidth}
\textbf{Dataset} & \textbf{German} & \textbf{Recidivism} & \textbf{Math} & \textbf{Por}\\
 \Xhline{1.5\arrayrulewidth}
\# Nodes & 1,000 & 18,876 & 649 & 649 \\
\# Features & 27 & 18 & 33 & 33\\
Sens. & Gender & Race & Gender & Gender\\
Label & Credit Risk & Recidivism & Grade & Grade\\
\Xhline{2\arrayrulewidth}
\end{tabular}
\vskip -1.25ex
\label{tab-dataset}
\end{table}

\subsection{Baseline Details}
\label{sec-baselines}
In Table~\ref{tab-main}, we compare our method CFA with baseline methods Reduction and Reweight which are designed to mitigate the traditional result-oriented bias. In this paper, we use the statistical parity as the fairness constraint.
\begin{itemize}
    \item \textit{Reduction}\footnote{\url{https://fairlearn.org/v0.5.0/api_reference/fairlearn.reductions.html}}~\cite{reductions} : It reduces the fairness problem to a cost-sensitive classification problem and trains a randomized classifier under certain fairness constraints.
    \item \textit{Reweight}\footnote{\url{https://github.com/google-research/google-research/tree/master/label_bias}}~\cite{reweight}: It reduces bias by re-weighting the data points during training process.
\end{itemize}

In Section~\ref{sec. graph}, we explore the metrics and algorithm in graph domain where we compare the methods and its augmented version with the explanation fairness loss including NIFTY and FairVGNN:
\begin{itemize}
    \item \textit{NIFTY}\footnote{\url{https://github.com/chirag126/nifty}}~\cite{fairvgnn}: It maximizes the similarity of node representations learned from the original graph and its variants to learns fair representations which are invariant to sensitive features.
    \item \textit{FairVGNN}\footnote{\url{https://github.com/yuwvandy/fairvgnn}}~\cite{fairvgnn}: It learns a mask to prevent sensitive feature leakage from sensitive-correlated feature channels and therefore alleviate the discrimination in the learned representations.
\end{itemize}

\subsection{Hyperparameters Setting}
\label{sec-hyper_parameters}
\subsubsection{(1) Grid search range}
The grid search range for each method is provided below:
\begin{itemize}
    \item MLP: learning rate $\{1e^{-2}\}$, weight decay $\{1e^{-3}, 1e^{-4}, 1e^{-5}\}$, dropout $\{0.1, 0.3, 0.5\}$.
    
    \item Reduction: regularization coefficient $\{0.001, 0.01, 0.1, 1, 10, 100\}$.
    
    \item Reweight: regularization coefficient $\{-1.0, -0.5, ..., 2.5, 3.0\}$.
    
    \item CFA: same hyperparameter space with MLP and regularization coefficient $\{0, 0.001, 0.01, 0.1, 1, 10\}$.
    
    \item NIFTY: we follow the same setting in the experiment for some hyperparameters: the number of hidden unit 16, learning rate $\{1e^{-3}\}$, project hidden unit 16, weight decay $\{1e^{-5}\}$, drop edge rate $0.001$, drop feature rate $0.1$ and tune dropout $\{0.1, 0.3, 0.5\}$, regularization coefficient $\{0.0, 0.2, 0.4, 0.6, 0.8, 1.0\}$.
    
    \item FairVGNN: we apply the best setting reported in the paper for each dataset and tune regularization in range $\{0.0, 0.1, ..., 0.9, 1.0\}$.

\end{itemize}

\subsubsection{Best Hyperparameters}
The best hyperparameters for each method after grid search are shown here where the evaluation criterion is $\text{Score}=(\text{AUC}+\text{F1}+\text{Acc})/3.0-(\Delta_{\text{SP}}+\Delta_{\text{EO}})/2.0-(\Delta_{\text{REF}}+\Delta_{\text{VEF}})/2.0$.
\begin{itemize}
    \item MLP: (1) German: [[0.01, 0.0001, 0.3], [0.01, 0.001, 0.3], [0.01, 0.001, 0.5], [0.01, 0.001, 0.1], [0.01, 1e-05, 0.1]]; (2) Recidivism: [[0.01, 0.001, 0.1], [0.01, 0.0001, 0.5], [0.01, 0.001, 0.3], [0.01, 0.001, 0.1], [0.01, 0.001, 0.1]]; (3) Math: [[0.01, 1e-05, 0.3], [0.01, 0.001, 0.5], [0.01, 1e-05, 0.5], [0.01, 1e-05, 0.1], [0.01, 0.001, 0.5]]; (4) Por: [[0.01, 0.0001, 0.5], [0.01, 0.0001, 0.5], [0.01, 1e-05, 0.5], [0.01, 0.001, 0.5], [0.01, 0.0001, 0.5]].
    
     \item Reduction: (1) German:[0.001, 0.001, 1.0, 1.0, 1.0]; (2) Recidivism: [10.0, 10.0, 10.0, 10.0, 10.0]; (3) Math: [10.0, 0.001, 10.0, 10.0, 10.0]; (4) Por: [0.001, 0.001, 0.001, 0.001, 0.001].
    
    \item Reweight: (1) German: [2.0, 2.0, 2.0, 2.0, 2.0]; (2) Recidivism: [3.0, 3.0, 3.0, 3.0, 3.0]; (3) Math: [0.5, 0.5, 0.0, 0.0, 0.0]; (4) Por: [0.0, 0.0, 0.0, 0.0, 0.0].
    
    \item CFA: (1) German: [[0.01, 0.001, 0.5, 10.0], [0.01, 1e-05, 0.3, 10.0], [0.01, 0.001, 0.1, 10.0], [0.01, 0.0001, 0.5, 10.0], [0.01, 0.0001, 0.3, 10.0]]; (2) Recidivism: [[0.01, 1e-05, 0.5, 0.01], [0.01, 0.001, 0.3, 0.01], [0.01, 0.001, 0.1, 0.001], [0.01, 0.001, 0.3, 0.01], [0.01, 0.001, 0.3, 0.01]]; (3) Math: [[0.01, 0.001, 0.5, 0.1], [0.01, 0.001, 0.5, 0.1], [0.01, 0.001, 0.1, 0.1], [0.01, 0.0001, 0.5, 0.01], [0.01, 1e-05, 0.5, 0.1]]; (4) Por: [[0.01, 0.0001, 0.1, 0.01], [0.01, 0.0001, 0.3, 0.01], [0.01, 0.0001, 0.3, 0.001], [0.01, 0.0001, 0.1, 0.01], [0.01, 0.001, 0.3, 0.001]].

\end{itemize}

\balance 

\end{document}